\documentclass[a4paper]{article}

\usepackage{amsfonts}
\usepackage{amssymb}
\usepackage{latexsym}
\usepackage{graphicx}

\usepackage{clrscode}
\usepackage{dsfont}

\title{Evolutionary Optimization for Decision Making under Uncertainty}
\author{Ronald Hochreiter}
\date{May 2011}

\begin{document}

\maketitle

\begin{abstract}
\noindent Optimizing decision problems under uncertainty can be done using a variety of solution methods. Soft computing and heuristic approaches tend to be powerful for solving such problems. In this overview article, we survey Evolutionary Optimization techniques to solve Stochastic Programming problems - both for the single-stage and multi-stage case.
\end{abstract}


\section{Introduction}

There are many approaches towards solving decision problems under uncertainty. Throughout this paper, we will focus on solving Stochastic Programming problems - both the classical single-stage as well as the multi-stage case. This article intends to provide ideas on how to use Evolutionary Optimization techniques to solve this class of optimization problems.

To summarize the concept of Stochastic Programming, we will refer to the summary of \cite{Hochreiter2008}: consider a deterministic decision optimization problem, where a decision maker aims at finding an optimal (numerical) decision $x \in \mathds{R}^n$ by minimizing a deterministic cost function $f(\cdot)$ (or by maximizing a profit function respectively) given a set $\mathcal{X}$ of constraints, which generally consists of various physical, organizational, and regulatory restrictions. The mathematical formulation of this problem can be simplified to the formulation shown in Equ. (\ref{equ:detprog}).
\begin{eqnarray}
\begin{array}{ll}
\mbox{optimize } x: & f(x) \\
\mbox{subject to} & x \in \mathcal{X}.
\end{array}
\label{equ:detprog}
\end{eqnarray}

During the 1950s Stochastic Programming was initiated by Dantzig \cite{Dantzig1955} and Beale \cite{Beale1955}. The idea is to replace deterministic parameters by probability distributions on some probability space $(\Omega, \mathcal{F}, \mathds{P})$, which will be denoted by $\Xi$ in the following, and to optimize a stochastic cost (or profit) function $f(\cdot, \cdot)$ over some probability functional $\mathds{F}$. A common choice regarding this functional is the expectation $\mathds{E}$. As Rockafellar \cite{Rockafellar2007} points out, expectations are only suitable for situations where the interest lies in long-range operation, and stochastic ups and downs can safely average out, which is not the case e.g. for managing financial risks. The recent progress of unifying probabilistic risk measures, as presented in the seminal paper by Artzner et al. \cite{ArtznerEtAl1999} on coherent risk measures, motivated for using probability functionals based on risk measures. See the book \cite{PflugRoemisch2007} for 
more details on modeling, measuring, and managing risk for this class of optimization applications. A different view on integration of risk measures using the concept of deviation measures is shown in \cite{RockafellarEtAl2006}. In summary, the resulting mathematical meta-formulation of a stochastic program for arbitrary probability functionals is shown in Equ. (\ref{equ:stochprog}).
\begin{eqnarray}
\begin{array}{ll}
\mbox{optimize } x: & \mathds{F}(f(x, \Xi)) \\
\mbox{subject to} & (x, \Xi) \in \mathcal{X}.
\end{array}
\label{equ:stochprog}
\end{eqnarray}

However, a concrete reformulation of this meta-model into some model, which can be solved with a numerical optimizer depends to a high degree on the chosen probability functional, as well as on the structure of the underlying probability space. 

The interested reader is referred to \cite{RuszczynskiShapiro2003} for a theoretical overview of the area of Stochastic Programming, and to \cite{WallaceZiemba2005} for Stochastic Programming languages, environments, and applications. Interestingly, evolutionary approaches have not been applied to a wide range of real stochastic programming problems so far, only scarce examples are available, e.g. recent works in the field of chemical batch processing, see \cite{UrselmannEtAl2007} and \cite{TillEtAl2007}. 

This paper is organized as follows. Section \ref{sec:evo} briefly describes the heuristic (soft computing) approach, which was chosen to solve the stochastic programs. Section \ref{sec:single} and Section \ref{sec:multi} consider the single-stage and the multi-stage case respectively. In both Sections the optimization as well as the scenario generation aspect will be discussed. Section \ref{sec:conclusion} concludes the paper.

\section{Evolutionary stochastic optimization}
\label{sec:evo}

All stochastic optimization problems throughout this paper will be solved by adapting a standard Evolutionary Optimization algorithm, e.g. as surveyed by \cite{BlumRoli2003} and summarized below, to handle the stochastic meta-model directly without complex reformulations. 

\begin{codebox}
\zi $P \gets GenerateInitialPopulation$
\zi $Evaluate(P)$
\zi \kw{while} termination conditions not met \kw{do}
\zi \> $P' \gets Recombine(P)$ 
\zi \> $P'' \gets Mutate(P')$ 
\zi \> $Evaluate(P'')$ 
\zi \> $P \gets Select(P \cup P'')$
\zi \kw{end while} 
\end{codebox}

As mentioned above, this meta-formulation of Evolutionary Optimization can be applied to the meta-formulation of Stochastic Programming shown in Equ. (\ref{equ:stochprog}) in different ways. We consider the case of finite, discrete scenario sets in this paper. This is not a serious restriction as many real-world problems can be satisfyingly described with such sets. In addition, analytical solutions and continuous descriptions of the uncertainty often imply a gross underestimation of the complexity of the respective problem.

\section{Single-stage stochastic optimization}
\label{sec:single}

In single-stage stochastic optimization, the input is given by a multi-variate probability distribution - a discrete one in our special application-related approach. In the following, issues raised by using Evolutionary Optimization techniques are shown using the well-known application of financial risk-return portfolio selection. In a second step, scenario generation for the single-stage case is boiled down to being solvable using an evolutionary approach.

\subsection{Single-stage optimization: Portfolio selection}

An ideal application example is financial risk-return portfolio optimization. For various reasons, this optimization problem naturally fits into the general problem structure usually handled with evolutionary techniques. Therefore, evolutionary approaches have been successfully applied to different classes of portfolio optimization problems, see e.g. \cite{StreichertEtAl03}, \cite{SchlottmannEtAl05}, \cite{SubbuEtAl2005}, \cite{GomezEtAl2006}, \cite{LinEtAl2005}, as well as the references therein, or refer to \cite{Maringer2005}. Some of the approaches mention the field of Stochastic Programming explicitly, e.g. \cite{Tokoro2001}, and \cite{WangEtAl2006}. 

The classical bi-criteria portfolio optimization problem based on Markowitz \cite{Markowitz1952} can be summarized as follows: An investor has to choose a portfolio from a set of (financial) assets $\mathcal{A}$ with finite cardinality $a = \vert \mathcal{A} \vert$ to invest her available budget. The bi-criteria problem stems from the fact that the investor aims at maximizing her return while aiming at minimizing the risk of the chosen portfolio at the same time. 

In the single-stage stochastic programming case, the underlying uncertainty is represented as a multi-variate probability distribution on the respective probability space. The distribution may either be continuous or discrete. The discretized probability distribution, which is used to compute the optimal decision is called {\em scenario set}. This discrete set of scenarios $\mathcal{S}$ has finite cardinality $s = \vert \mathcal{S} \vert$, where each $s_i$ is equipped with a non-negative probability $p_i \geq 0$, and $\sum_{j=1}^s p_j = 1$. From the financial market viewpoint, each scenario contains one possible set of joint future returns of all $a$ assets under consideration for the portfolio. Using the terminology of Markowitz, each scenario contains the discounted anticipated return of each asset.

Let $x \in \mathds{R}^a$ be some portfolio. Without loss of generality, we use budget normalization, i.e. $\sum_{a \in \mathcal{A}} x_a = 1$. Each component $x_i$ of the portfolio vector denotes the fraction of the available budget $B$ invested into the respective asset $i$. We may now rewrite the scenario set $\mathcal{S}$ as a matrix $S$ to calculate the discrete Profit \& Loss (P\&L) distribution $\ell$ for some portfolio $x$, which is simply the cross product $\ell_x = \big < x, S \big >$. We will denote $\ell$ the loss distribution in the following. Finally, let $x_\rho^* \in \mathds{R}^a$ denote the optimal portfolio given some risk measure $\rho$ and $\ell_\rho^*$ denote the respective $\rho$-optimal discrete loss distribution. 

While this model clearly represents a multi-criteria optimization model, we apply a criteria-weighted model, where an additional risk-aversion parameter $\kappa$ is defined, i.e.
\begin{equation}
\begin{array}{ll}
\mbox{maximize } x: & \mathds{E}(\ell_x) - \kappa \rho(\ell_x), \\ 
\mbox{subject to } & x \in \mathcal{X}.
\end{array}
\label{equ:popt2}
\end{equation}

An important fact is that this problem already shows the issue that one really has to care about the most is the genotype-phenotype encoding. The most straightforward approach to design a usable genotype-phenotype representation of the portfolio selection problem is to use a vector of real values between $0$ and $1$ and normalize the resulting vector to the sum of the available budget, i.e. in the most simple case to a budget of $1$. However, it can be shown that this approach rarely leads to stable solutions, i.e. two runs almost never converge to the same portfolio composition, see also \cite{Hochreiter2010b}. 

Hence, we need a special genetic encoding of a portfolio, see \cite{StreichertEtAl03}. Each gene thereby consists of two parts: One that determines the amount of budget to be distributed to each selected asset and one part which determines in which assets to invest. The first part $g_1$ consists of a predefined number $b$ of real values between $0$ and $1$ and the second part $g_2$ is encoded as a bit-string of the size of the amount of assets.
\\ \par
\noindent {\bf Example (from \cite{Hochreiter2010b}).} Let us define a bucket size of $b=10$ and consider that want to select the optimal portfolio out of $a=5$ assets. A random chromosome with a fixed number of $3$ asset picks may then consist of the following two parts $g_1$ and $g_2$. 
$$g_1 = (0.4893, 0.3377, 0.9001, 0.3692, 0.1112, 0.7803, 0.3897, 0.2417, 0.4039, 0.0965)$$
$$g_2 = (1, 1, 0, 0, 1)$$
If we re-map these two parts, we receive the following portfolio $x$
$$x = (0.3, 0.5, 0, 0, 0.2),$$
which is a valid portfolio composition, and can be used to calculate the loss function and to  conduct a full evaluation of the respective portfolio optimization formulation within the evolutionary optimization process.

With this Evolutionary Optimization strategy a full-fledged portfolio optimization tool for a plethora of risk measures can be created, as the evolutionary approach only needs to evaluate the risk and not directly optimize over the complete (probably non-convex) problem. Furthermore, all kinds of constraints, even non-convex ones, may be included. Constraint handling can be easily added using a penalty structure within the objective function evaluation.

\subsection{Single-stage scenario generation}

The problem of single-stage scenario generation, i.e. an optimal approximation of a multi-variate probability distribution can be done via various sampling as well as clustering techniques. Optimal single-stage approximation have been done by using e.g. Principal Component Analysis (see e.g. \cite{TopaloglouEtAl2002}), K-Means clustering (see e.g. \cite{HochreiterPflug2007}), or moment matching (\cite{HoylandWallace2001}). The result is directly affected by the chosen approximation method - both a distance and a heuristic needs to be specified.

Of course, this problem can be solved with Evolutionary Optimization techniques as well. Consider the following example (from \cite{Hochreiter2010}) of a full-fledged scenario generation procedure: let's assume that we do have $10$ input scenarios (asset returns), each equipped with the same probability $p=0.1$, which might be the output of some sophisticated asset price sampling procedure, e.g.
$$(0.017,-0.023,-0.008,-0.022,-0.019,0.024,0.016,-0.006,0.032,-0.023).$$
We want to separate those values optimally into $2$ clusters, which then represent our output scenarios and take a random chromosome, which might look as follows:
$$(0.4387,0.3816,0.7655,0.7952,0.1869,0.4898,0.4456,0.6463,0.7094,0.7547)$$
If we map this vector to represent $2$ centers we obtain: $(1,1,2,2,1,1,1,2,2,2)$. Now we need to calculate a center value, e.g. the mean, and have to calculate the distance for each value of each cluster to its center, e.g. we obtain center means $(0.0032,-0.0055)$, which represent the resulting scenarios, each with a probability of $0.5$. The $l_1$ distance for each cluster is $(0.0975, 0.0750)$, so the objective function value is $0.1725$. Now flip-mutate chromosome $9$, i.e. $(1-0.7094)=0.2906$, such that input scenario $9$ (return = $0.032$) will now be part of cluster $1$ instead of cluster $2$. We obtain new scenarios $(0.0080, -0.0149)$ with probabilities $(0.6, 0.4)$. The objective function value is $0.1475$ (or $0.1646$ if you weight the distances with the corresponding output scenario probability), i.e. this mutation led to a better objective value.

\subsection{Wrapping up the single-stage case}

In this Section we have shown that Evolutionary Optimization can be conveniently be applied to single-stage stochastic optimization problems - both to the solution of decision models as well as to the creation of scenario sets, i.e. for the process of scenario generation. However, one has to be careful with choosing an appropriate genotype-phenotype representation, as the most straightforward approach might not lead to a stable solution. 

\section{Multi-stage stochastic optimization}
\label{sec:multi}

To summarize the case of multi-stage stochastic optimization, we base the following summary on \cite{Hochreiter2009}: we consider algorithmic issues of the computation of multi-stage scenario-based stochastic decision processes for decision optimization models under uncertainty. We consider that given a multi-stage stochastic programming problem specific discrete-time stochastic process on the decision horizon $t = 1, \ldots, T$, a decision maker observes the realization of this random process $\xi_t$ at each decision stage $t$, and takes a decision $x_t$ based on all observed values up to $t$ ($\xi_1, \ldots, \xi_t$). Let there now be a sequence of decisions $x_1, \ldots, x_T$. At the terminal stage $T$ we observe a sequence of decisions $x = (x_1, \ldots, x_T)$ with realizations $\xi = (\xi_1, \ldots, \xi_T)$, which lead to cost $f(x, \xi)$ (or likewise profit). The stochastic optimization task is to find the sequence of decisions $x(\xi)$, which minimizes some probability functional (most commonly the 
expectation, a risk measure, or a combination of these two) of the respective cost function $f(x(\xi), \xi)$ - see e.g. \cite{EichhornRoemisch2005} for a classification of risk measures in this context. 

We consider the multi-stage case in such a way that there is at least one intermediary stage between root and terminal stage, i.e. $T>2$. Unfortunately, the topic of scenario generation is not sufficiently treated in most text books on stochastic programming. Let us consider multi-stage stochastic programming problems as defined in Equ.~(\ref{equ:mssp}).
\begin{eqnarray}
\begin{array}{lll}
\mbox{minimize } x: & \mathds{F} \big ( f(x(\xi), \xi) \big ) & \\
\mbox{subject to} & (x(\xi), \xi) \in \mathcal{X} & \\
& x \in \mathcal{N} &
\end{array}
\label{equ:mssp}
\end{eqnarray}
The multi-variate, multi-stage stochastic process $\xi$ describes the future uncertainty, i.e. the subjective part of the stochastic program, and the constraint set $\mathcal{X}$ defines feasible combinations of $x$ and $\xi$. This constraint set is used to model the underlying real-world decision problem. Furthermore, a set of non-anticipativity constraints $\mathcal{N}$, consisting of functions $\xi \mapsto x$ which make sure that $x_t$ is only based on realizations up to stage $t$ $(\xi_1, \ldots, \xi_t)$, is necessary. To solve multi-stage programs with numerical optimization solvers, the underlying stochastic process has to be discretized into a scenario tree, and this scenario tree approximation will inherently fulfill these non-anticipativity constraints.

Two issues negatively affect the application of multi-stage stochastic programs for real-world decision problems: First, modeling the underlying decision problem is a non-trivial task. Multi-stage optimization models and stochastic scenario models require stable scenario tree handling procedures, which are considered to be too cumbersome to be applied to real-world applications, and the communication of tree-based models to non-experts is complicated. Secondly, modeling the underlying uncertainty is complex and messy. As discussed above, a good discrete-time, discrete-space scenario tree approximation of the underlying stochastic process has to be generated in order to numerically compute a solution and the quality of the scenario model severely affects the quality of the solution.

\subsection{Solving the scenario tree optimization problem}
\label{sec:tree}

The quality of the scenario tree severely affects the quality of the solution of the multi-stage stochastic decision model, such that any approximation should be done in consideration of optimality criteria, i.e. before a stochastic optimization model is solved, a scenario optimization problem has to be solved independently of the optimization model. It should be noted, that there are also scenario generation approaches where the generation is not decoupled from the optimization procedure, see e.g. \cite{CaseySen2005}. A major drawback is that these methods are often limited to a certain restricted set of models, and cannot be generalized easily.

In the context of separated scenario optimization, optimality can be defined as the minimization of the distance between the original (continuous or highly discrete) stochastic process and the approximated scenario tree. Choosing an appropriate distance may be based on subjective taste, e.g. Moment Matching as proposed by \cite{HoylandWallace2001}, selected due to theoretical stability considerations (see \cite{RachevRomisch2002} and \cite{HeitschEtAl2006}), which leads to probability metric minimization problems as shown by \cite{Pflug2001} and \cite{DupacovaEtAl2003}, or it may be predetermined by chosen approximation method, e.g. by using different sampling schemes like QMC in \cite{PennanenKoivu2005} or RQMC in \cite{Koivu2005}, see also \cite{Pennanen2009}. It is important to remark that once the appropriate distance has been selected, an appropriate heuristic to approximate the chosen distance has to be applied, which affects the result significantly. The real algorithmic challenge of multi-stage 
scenario generation is caring about the tree structure while still minimizing the distance. Only in rare cases, this problem can be solved without heuristics. 

The following evolutionary approach has been presented by \cite{Hochreiter2010}. We assume that there is a finite set $\mathcal{S}$ of multi-stage, multi-variate scenario paths, which are sampled using the preferred scenario sampling engine selected by the decision taker. Stages will be denoted by $t = 1, \ldots, T$ where $t=1$ represents the (deterministic) root stage (root node), and $T$ denotes the terminal stage. Therefore, the input consists of a scenario path matrix of size $\vert \mathcal{S} \vert \times (T-1)$. Furthermore, the desired number of nodes of the tree in each stage is required, i.e. a vector $n$ of size $(T-1)$.

We will focus on the uni-variate case. However, the extension to the multi-variate case does not pose any structural difficulties besides that a dimension-weighting function for calculating the total distance on which the optimality of the scenario tree approximation is based on has to be defined.

A crucial part in designing a multi-stage scenario tree generator based on evolutionary techniques is finding a scalable genotype representation of a tree, as this is always the case for Evolutionary Optimization approaches - specifically both in terms of the numbers of stages as well as the number of input scenarios. The approach surveyed here is using a real-valued vector in the range $[0,1]$ and mapping it to a scenario tree given the respective node format $n$. The length of the vector is equal to the number of input scenarios $s = \vert \mathcal{S} \vert$ plus the number of terminal nodes $n_T$. Thus, the presented algorithm is somewhat limited by the number of input scenarios. This means that input scenarios should not simply be a standard set of mindlessly sampled scenario paths, but rather a thoughtfully simulated view on the future uncertainty. This should not be seen as a drawback, as it draws attention to this often neglected part of the decision optimization process.

To map the real-valued vector to a scenario tree, which can be used for a subsequent stochastic optimization, two steps have to be fulfilled. First, the real-valued numbers are mapped to their respective node-set given the structure $n$ of the tree, and secondly, values have to be assigned to the nodes. It should be noted, that a random chromosome does not necessarily lead to a valid tree. This is the case if the number of mapped nodes is lower than the number of nodes necessary given by $n_t$ of the respective stage $t$. If an uniform random variable generator and a thoughtful node structure is used, which depends on the number of input scenarios, invalid trees should not appear frequently, and can be easily discarded if they do appear during the Evolutionary Optimization process.

We may now conveniently use the Evolutionary Optimization clustering strategy, which has been shown for creating single-stage scenarios above. This approach seems to be rather trivial for the single-stage case, but this simple approach leads to a powerful method for the tedious task of constructing multi-stage scenario trees for stochastic programming problems, because the nested probability structure of the stochastic process is implicitly generated.

\begin{figure}
\begin{center}
\begin{tabular}{cc}
\scalebox{0.3}{
\includegraphics{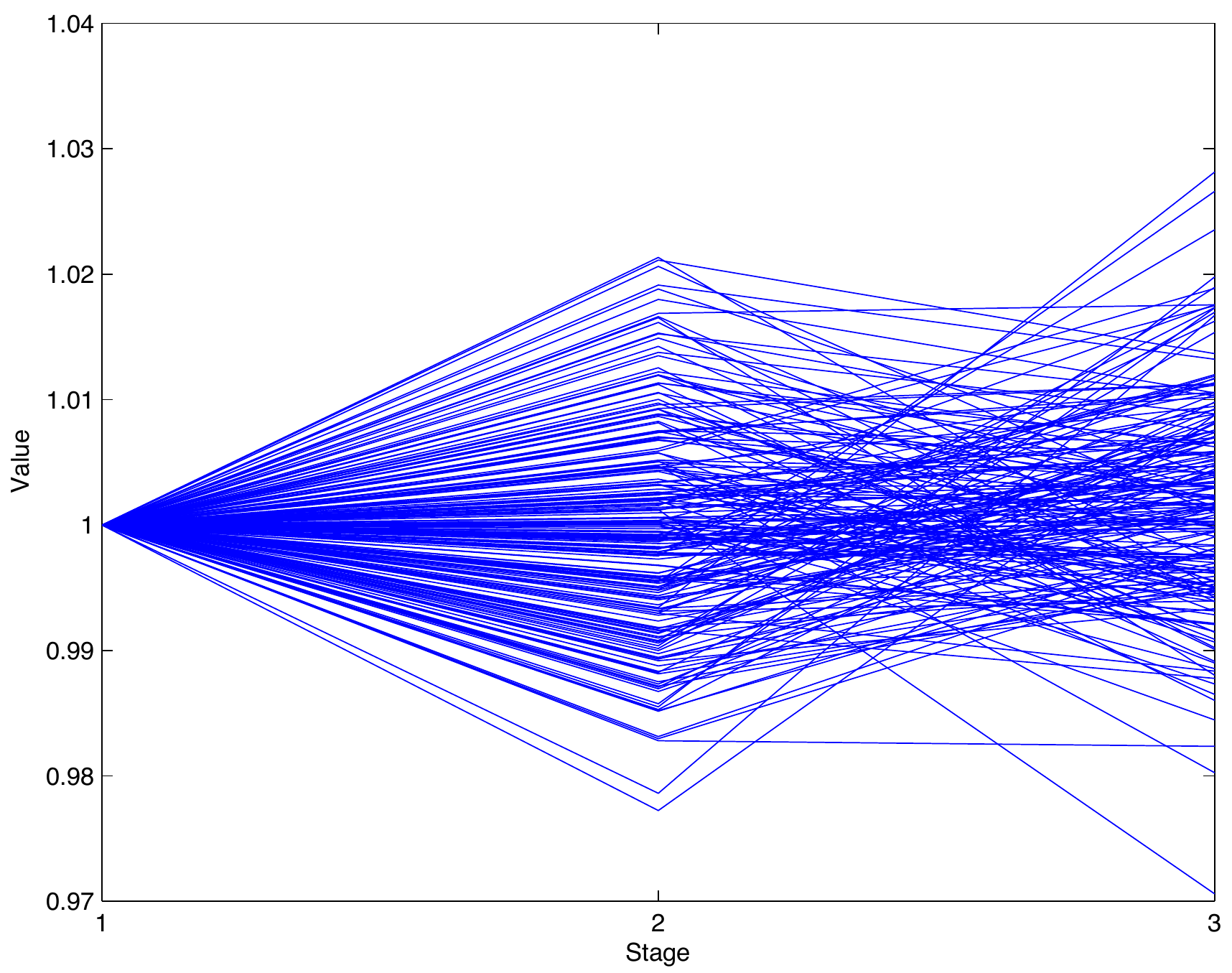}
}
&
\scalebox{0.3}{
\includegraphics{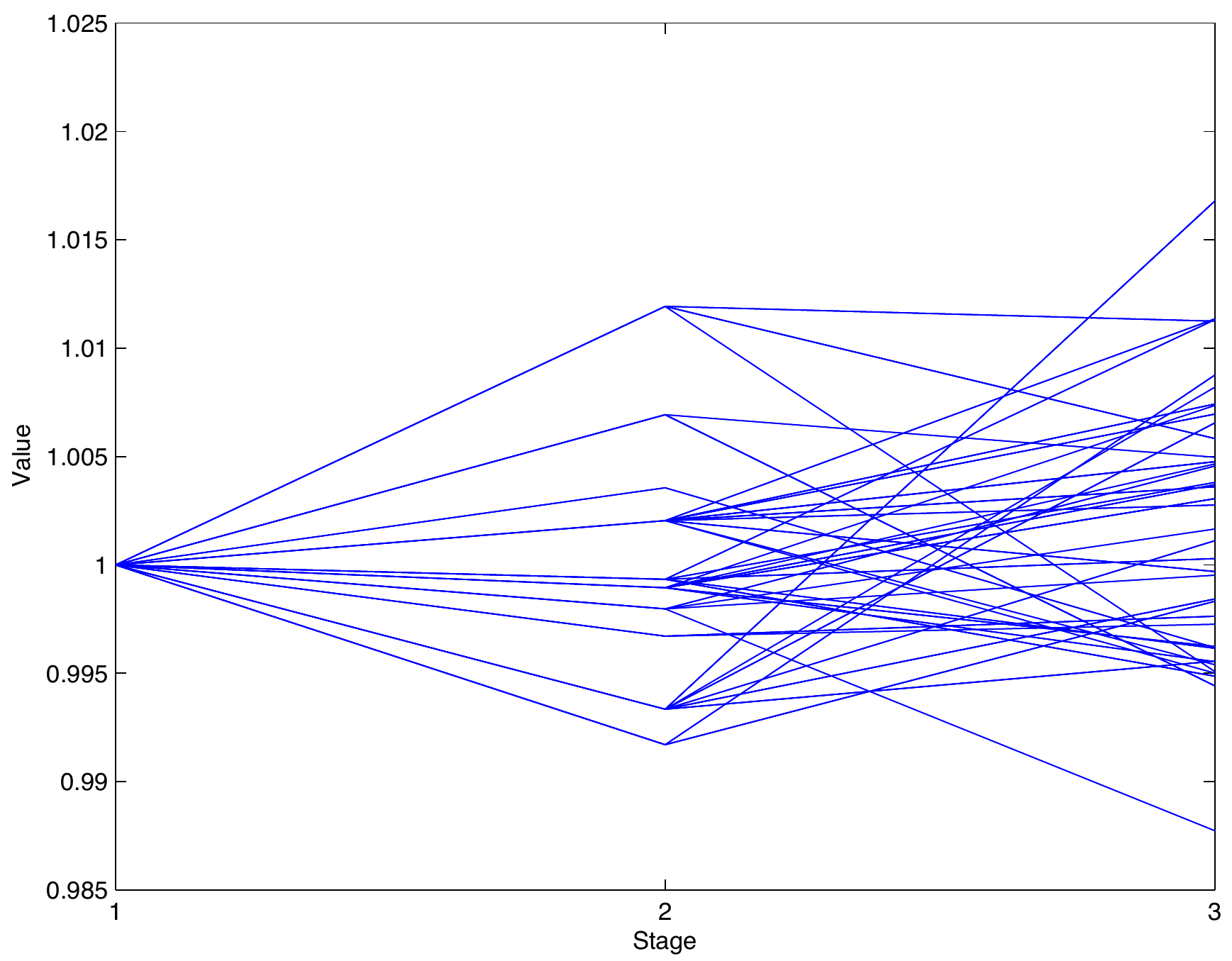}
}
\end{tabular}
\end{center}
\caption{Input scenarios $(n=200)$ (left) and a scenario tree with $n = [10,40]$ (right).}
\label{fig:treesize}
\end{figure}

For multi-stage trees, a crucial point is finding a representative value for the node-sets determined in the first step of the mapping. There exists a range of methods, which can be used for the determination of centers, i.e. mapping all values of a node-sets to one node. The distance of the approximation, which is used for the calculation of the objective function, will also be affected by this method. A straight-forward solution is to use the median of the values - see  \cite{Hochreiter2010} for other methods and their differences. Fig. \ref{fig:treesize} shows an example of the algorithm, i.e. $200$ sampled paths are used to create a multi-stage $[1,10,40]$ scenario tree. In summary, the complex task of generating multi-stage scenario trees can be handled conveniently by using an Evolutionary Optimization approach.

\subsection{Solving the multi-stage decision problem}
\label{sec:multiprob}

Solving a multi-stage stochastic programming problem with Evolutionary Optimization techniques is not straightforward. Especially when the underlying scenario tree is huge, because one would have to calculate an optimal decision on every node, such that the chromosome would simply be too large, i.e. for each node of the tree the respective amount of genes (related to the multi-variate structure of the decision problem) would have to be added to the chromosome. New simplification strategies have to be devised, which are out of the scope of this paper.

\section{Conclusion}
\label{sec:conclusion}

In this paper, several methods on how to apply Evolutionary Optimization methods to solve stochastic programs have been surveyed and outlined. It is shown that this class of heuristic optimization solvers can be applied to these problems conveniently. However, one has to take care of a few pitfalls during the creation of specific solution methods, which were discussed in this paper.
\\ \par
\noindent {\bf Acknowledgement:} The author is grateful to the MENDEL conference organizing committee for being invited to present this research to the soft computing community. 

\bibliographystyle{plain}
\bibliography{eofdmuu}

\end{document}